\pdfoutput=1
\documentclass[journal]{IEEEtran}
\usepackage{graphicx}
\usepackage{hhline}
\usepackage{slashbox}

\newcommand{\etal}{\textit{el al. }}
\usepackage{gensymb}
\usepackage{paralist}

\usepackage{tabularx,booktabs}
\usepackage{hyperref}
\hypersetup{
    colorlinks=true,
    linkcolor=blue,
    filecolor=magenta,      
    urlcolor=cyan,
}
\usepackage{color}
\definecolor{red}{rgb}{1, 0, 0}

\def\imwX#1#2{\includegraphics[clip,width=#2\textwidth]{img/#1.pdf}}
\newcommand{\tb}[3]{\setlength{\tabcolsep}{#2mm}\begin{tabular}{#1}#3\end{tabular}}

\def\imw_#1#2{\fbox{\includegraphics[clip,width=#2\textwidth]{#1.pdf}}}

\def\imw__#1#2#3{
\begin{minipage}[b]{#2\linewidth}
  \centering
  \centerline{\includegraphics[width=1\textwidth]{img/#1.pdf}}
\end{minipage}
}
\ifCLASSINFOpdf
\else
\fi
\usepackage{multirow}
\usepackage{array}
\usepackage{booktabs}
\usepackage{amssymb}
\usepackage{mathtools}

\begin{document}
\title{
Successive Embedding and Classification Loss for 
Aerial Image Classification
}

\author{Jiayun Wang, \textit{Student Member, IEEE},
        Patrick Virtue, \textit{Student Member, IEEE},
        and~Stella X. Yu, \textit{Member, IEEE}
\thanks{
The authors are with UC Berkeley and ICSI, 1947 Center St, Berkeley, CA, 94704.  Tel: +1 (510) 666-2900, Fax: +1 (510) 666-2956. E-mail: \{peterwg, virtue, stellayu\}@berkeley.edu.}
}

\markboth{ }%
{Wang \MakeLowercase{\textit{et al.}}: Successive Embedding and Classification Loss for Aerial Image Classification}

\maketitle

\begin{abstract}
 Deep neural networks can be effective means to automatically classify aerial images but is easy to overfit to the training data.
It is critical for trained neural networks to be robust to variations that exist between training and test environments.  To address the overfitting problem in aerial image classification, we consider the neural network as successive transformations of an input image into embedded feature representations and ultimately into a semantic class label, and train neural networks to optimize image representations in the embedded space in addition to optimizing the final classification score. We demonstrate that networks trained with this dual embedding and classification loss outperform networks with classification loss only. 
We also find that moving the embedding loss from commonly-used feature space to the classifier space, which is the space just before softmax nonlinearization, leads to the best classification performance for aerial images. Visualizations of the network's embedded representations reveal that the embedding loss encourages greater separation between target class clusters for both training and testing partitions of two aerial image classification benchmark datasets, MSTAR and AID. Our code is publicly available on GitHub \cite{githubrepo}.

\end{abstract}


\IEEEpeerreviewmaketitle

\section{Introduction}
\IEEEPARstart{U}{nderstanding} of the semantic content of aerial images has many real-world applications in geoscience and remote sensing \cite{area1, area2, area3}. In this paper, we focus on two aerial image classification scenarios:  target recognition of synthetic aperture radar (SAR) images, and classification of aerial scene images.
\begin{figure}[t!]
\centering
\includegraphics[width=0.48\textwidth]{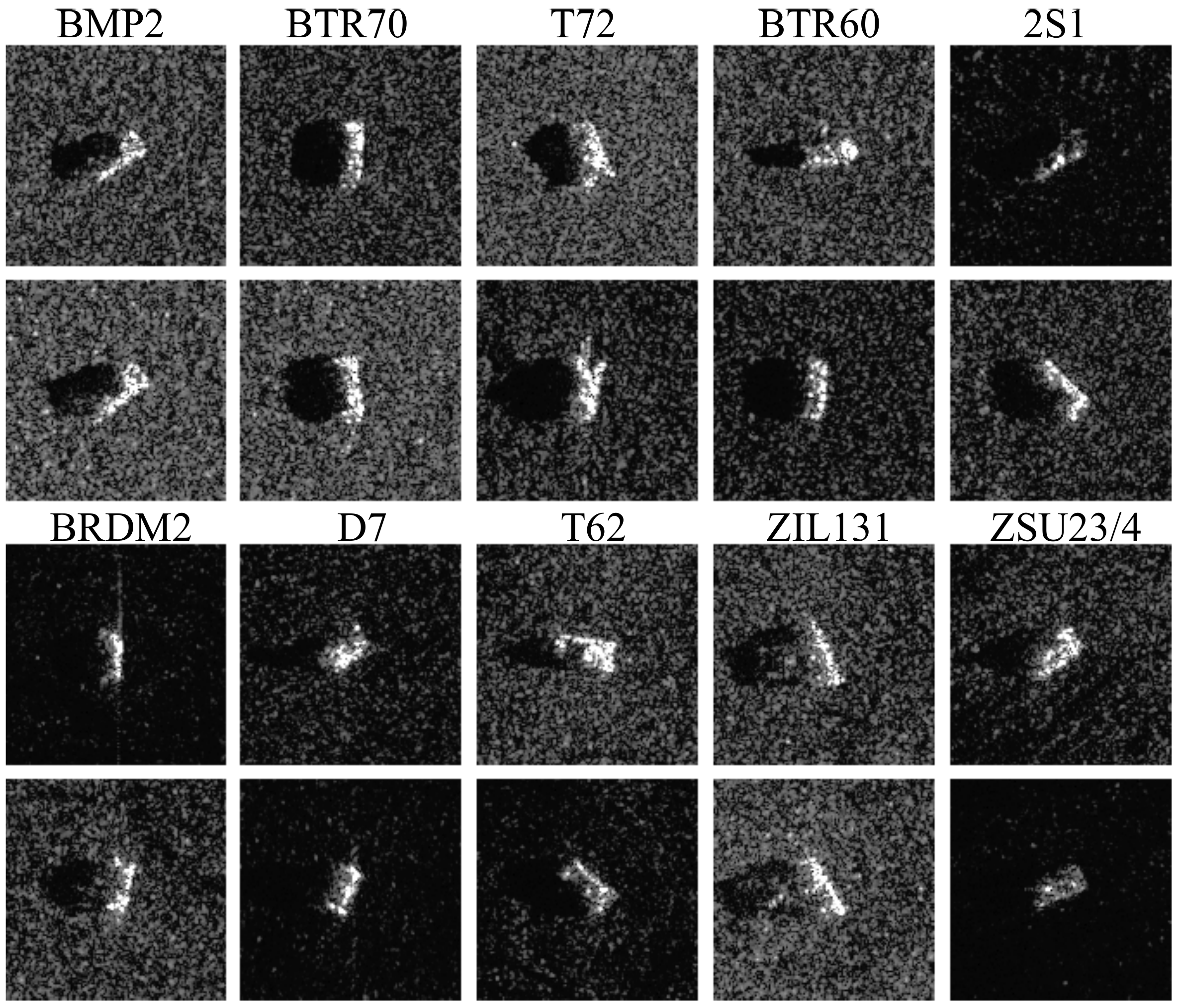}\\
\includegraphics[width=0.48\textwidth]{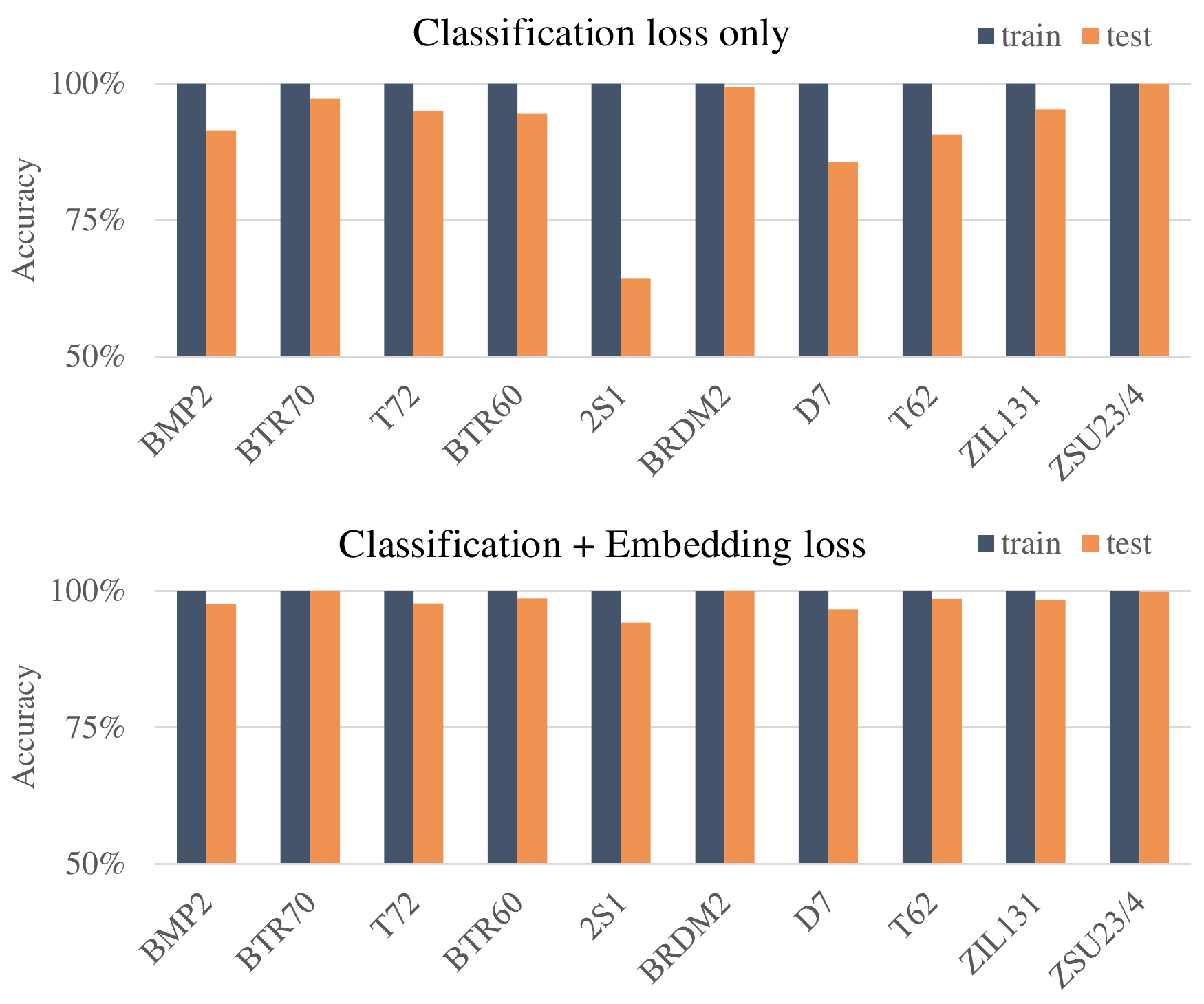}
\caption{Our goal is to reduce the overfitting problem in aerial image classification with a dual-loss criterion. We take target recognition in SAR images as an example.
{\bf Top}: Sample training (first and third row) and testing chips (second and fourth row) for each class of the MSTAR dataset \cite{ding}. We apply \textit{translation augmentation} to the data in our experiments.
A typical neural network target classifier fails on these test samples, whereas our new method can successfully classify them.
{\bf Bottom}: A deep neural network with only a classification loss (e.g. softmax loss) reaches 100\% training accuracy but only 91.4\% test accuracy. With our proposed dual classification and embedding loss, the modified network can achieve 98.2\% test accuracy, significantly reducing the gap between training and testing accuracies.
}
 \label{fig:main_results}
\end{figure}

\textit{Synthetic aperture radar} captures images of objects that are suitable for target classification, reconnaissance and surveillance, regardless of whether it is day or night.
Target recognition in SAR images has many potential applications in military and homeland security, such as friend and foe identification, battlefield
surveillance, and disaster relief.
Automatic target recognition (ATR) for SAR images often proceeds in two stages: detection and classification.
At the detection stage, candidate target patches are extracted from  SAR images; they may include not only targets of interest, but also false alarm clutter such as trees, cars, buildings, etc. 
These detected image regions, or \textit{chips}, are then sent to the classification stage, where a classification algorithm predicts the target class of each chip.  In this work, we focus on improving the chip classification accuracy.

\textit{Aerial images}, as significant data source for earth observation, enable the possibility to measure the earth surface with detail structures \cite{area3, cheng2015effective}.
Aerial scene classification aims at automatically assigning a semantic scene label to each image. As an important and challenging topic in remote sensing, it has received increasing attention recently \cite{area1, scene7, scene8, scene9} due to the large variety and complicated composition of land-cover types. 

Traditional aerial image classification methods rely on extracting informative features and then training a classifier to map those features to semantic class labels. For target recognition of SAR images, various features have been explored, including raw pixel intensity \cite{chen9}, magnitudes of the two dimensional (2-D) DFT coefficients \cite{chen10}, and a transformation of SAR images into polar coordinates followed by PCA compression \cite{chen3}. Associated feature classifiers include support vector machines \cite{chen9}, radial basis function networks \cite{chen10} and nearest neighbor approaches \cite{chen3}. Similarly, in aerial scene classification, the features for traditional classification methods includes the local structural texture descriptor \cite{risojevic2011aerial}, the orientation difference descriptor \cite{risojevic2012orientation}, and  multiscale features from the images with different spatial resolutions\cite{luo2013indexing}.

\begin{figure*}[t!]
\centering
\includegraphics[width=0.98\textwidth]{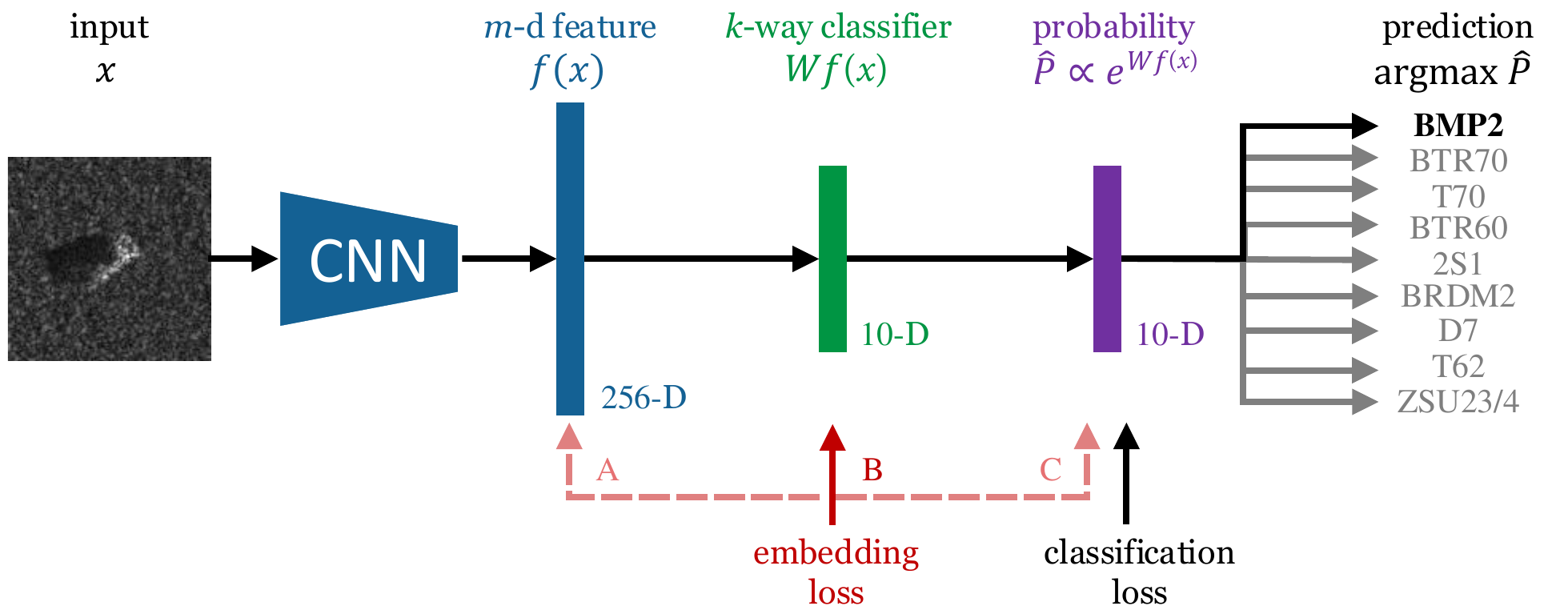}
\caption{A convolutional neural network (CNN) transforms the input image, $x$, into a series of embedded feature representations. We depict the final three embedding spaces in our network for SAR-ATR: the 256-D \textbf{feature space} $f(x)$, the 10-D \textbf{classifier space} from a learned linear projection of the feature layer $Wf(x)$, and the 10-D \textbf{probability space} after the nonlinear softmax normalization $\hat{P}$. Our experimental results show that applying the embedding loss to the classifier space $Wf(x)$ (B) is more effective than applying it to the other two spaces (A, C).}
 \label{fig:intro_2}
\end{figure*}

\def\widthA{0.47}
\def\widthB{0}
\def\widthC{0.32}
\def\widthD{0.5}


\begin{figure*}[ht]
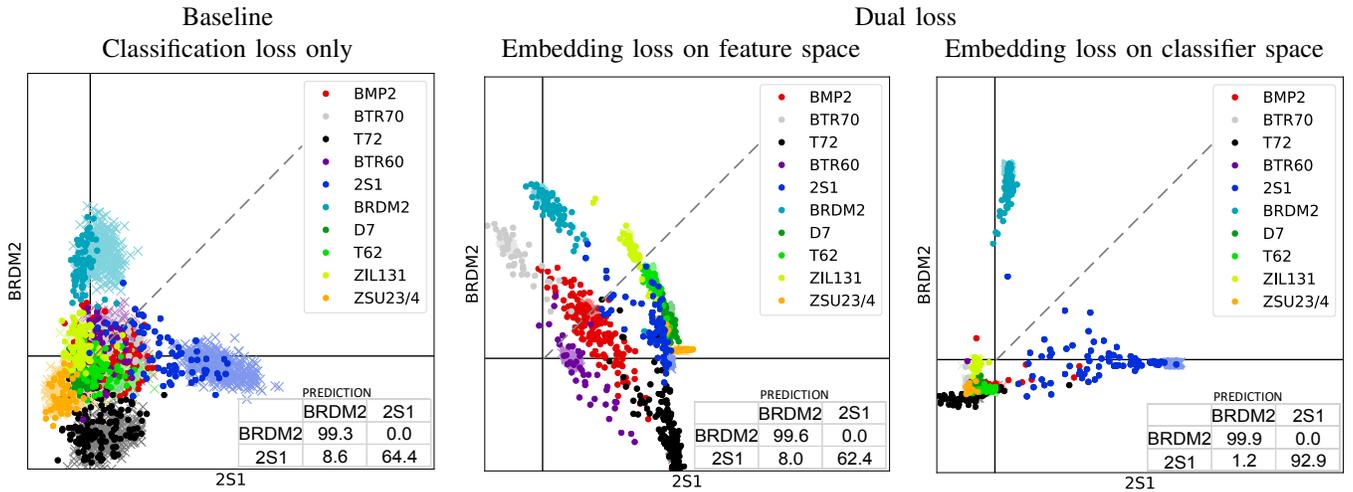

\centering
\tb{@{}ccc@{}}{0.5}{
{Baseline} & \multicolumn{2}{c}{Dual loss} \\
{Classification loss only} & {Embedding loss on feature space} & Embedding loss on classifier space \\
	\imw__{orth_base}{\widthC}{\footnotesize } &
	\imw__{orth_fc6}{\widthC}{\footnotesize } &
	\imw__{orth_fc7}{\widthC}{\footnotesize } 
}
\caption{Orthogonal projection of 10-D classifier space onto two target axes (BRDM2 and 2S1). Bold dots represent a subset of test points, while faint X's represent a subset of training points. The table on the lower right shows the confusion matrix for the two targets. The softmax loss only classifier (left) leads to large intra-class variance and little to no separation between classes. Adding an embedding loss on the feature space (center) reduces the intra-class variance but there is still limited separation between BRDM2 and 2S1 test point in classifier space. Moving the contrastive loss to the classifier space (right), the clusters of training points are pushed farther along their respective axes, helping to reduce the classification errors at test time.}
\label{fig:ortho}
\end{figure*}

Deep learning methods have recently been used for aerial image classification, integrating feature extraction and classification into a single neural network \cite{chen,chen7,ding,linSAR}. Some researchers focused on augmenting data to alleviate overfitting caused by insufficient image samples. Pei et al. \cite{pei2018sar} proposed a flexible method to generate SAR data from different camera view, which can provide a large amount of augmented data for network training. They also explored more optimized multiview networks for SAR-ATR \cite{pei2018multi, pei2018multiview}. Shang et al. \cite{sar-mnet} proposed a deep convolutional neural network architecture, which attempted to remember samples’ spatial features, to alleviate the problems caused by insufficient SAR image samples. 

Other deep learning works explored novel network architecture for aerial image classification. 
Furukawa \cite{furukawa2018deep} proposed a verification support network (VersNet) to perform all stages of SAR-ATR end-to-end. Peng et al. \cite{spp-sar} proposed a CNN architecture with spatial pyramid pooling (SPP) which utilized feature maps from finer to coarser levels to aggregate local features of SAR images. Xue et al. \cite{xue2018target} applied heterogeneous CNN ensemble on SAR-ATR and observed improved recognition accuracy. Vaduva et al. \cite{vaduva2012deep} introduced deep learning in high resolution remote sensing image scene classification. Mnih et al. \cite{mnih2012learning} considered the effect of noisy labels for aerial scene data in deep learning. Chen et al. 
\cite{chen2014deep} introduced a hybrid framework based on stacked auto-encoders for aerial scene classification. Hu et al. \cite{hu2015transferring} studied methods to transfer pre-trained CNN features for high-resolution remote sensing (HRRS) scene classification. 

Despite the appreciable achievements that these deep models have achieved in aerial image classification, the complexity of the real-world aerial images still presents challenging differences between the training and different test scenarios, including variations in the depression angle, remote sensors, landscape, target articulation, and target configuration. Therefore, we hope to propose a method to alleviate the variations between training and testing scenarios (Fig. \ref{fig:seen_unseen}).

The typical overfitting problem in SAR target recognition, for example, is zero training error (Fig. \ref{fig:main_results}), i.e., there is no room left for improving the learning process on the training set, yet there remains a large classification error (e.g. 9\%) on the test set.

We argue that increasing the separation of features between classes should alleviate the overfitting problem. A larger separation allows a larger margin of errors when the test set deviates from the training set, thereby reducing the gap between training and testing accuracies.

Classification using deep learning methods can be understood as successive transformation of feature representations, where raw intensities in the pixel space ultimately get transformed into labels in the semantic space. All the representations, including intermediate ones, can be considered as embeddings of the input images. When training a classification network, the spatial clustering of learned points in the embedding space become more indicative of individual semantic classes. 

We propose to augment the network training objective by replacing the standard single classification loss, e.g. softmax cross entropy loss, with a dual embedding and classification loss. The embedding loss aims at enlarging the distance between different semantic classes in the embedded feature space. In this work, we evaluate two embedding losses, contrastive loss \cite{bromley1994signature} and center loss \cite{wen2016discriminative}. With the additional embedding loss, training performance will still be high, but now if the test data vary away from the training data, the larger separation in the embedded space reduces the chance for misclassification.

The idea of combining a feature embedding loss with a target classification loss has been explored in the person re-identification (Person-ReID) task. This task, however, is not performing target identity classification, but determining whether two photos are of the same {\it unknown} person or not \cite{sun2014deep,wen2016discriminative,schroff2015facenet,p2s,adamargin}.  The feature embedding loss is a form of metric learning extending properties from samples on known identities to unknown identities, and it has thus developed naturally in a Person-ReID classification formulation.

In this work, we explore adding a feature embedding loss to the usual classification loss in the recognition of {\it known} identities.  We desire the learned features to be maximally separated from different classes, while supporting perfect classification of individual classes on the training set.  Consequently, the deep neural net becomes more robust to appearance deviation of unseen new instances of known categories in the test set.

In addition to the novel application of the dual embedding and classification loss to aerial image classification, we further improve the algorithm by imposing the embedding loss at a later representation stage, the classifier space\footnote{Classifier space refers to where the features are linearly mapped to the same dimension as the number of output classes and yet before the nonlinear softmax layer.}, instead of the feature space that is commonly done in the Person-ReID literature (Fig. \ref{fig:intro_2}). Visualizations  of  the network’s learned representations reveal  that  the embedding  loss on classifier space encourages  greater  separat0ion  between  target class cluster.

We summarize our paper as follows: \begin{inparaenum} [\bfseries (1)] \item
  We add the embedding loss to train the deep convolutional neural network (CNN), in order to reduce the overfitting problem of the usual single classification loss alone and to learn more discriminative feature representations and improve SAR target and aerial scene recognition accuracies. \item  We demonstrate that it is better to apply the additional embedding loss to the classifier space, instead of applying it to the commonly-used feature space \cite{cheng2018deep, sun2014deep}. We illustrate this by providing further insights through visualization of the learned discriminative feature space. 
  
\end{inparaenum}

\section{Proposed Method}
\label{sec:method}

In this section, we first present an overview of the proposed method,  then describe the details of our dual loss  in aerial image classification, and finally demonstrate that it is better to apply the embedding loss to the classifier space through discussions on the options for where to place the additional loss. 


\subsection{Overview of the proposed method}

As depicted in Fig. \ref{fig:intro_2}, a convolutional neural network (CNN) transforms the input image $x$ into a series of embedded feature representations. Take our CNN architecture used for SAR-ATR as an example, we consider the final three embedding spaces (with detailed definition in the following paragraph) in our network: the 256-D \textbf{feature space} $f(x)$, the 10-D \textbf{classifier space} from linear projection of the feature layer $Wf(x)$, and the 10-D \textbf{probability space} after the nonlinear softmax normalization $\hat{P}$.

We now formally define three specific embedding spaces in our network: The \textit{feature space} refers to the last network layer representation before the network maps it to the \textit{classifier space} where the dimension is the same as the output -- the total number of classes $k$.  The \textit{probability space} is the result 
 of normalizing the $k$-D classifier space via the softmax operation.  Specifically:
\begin{align}
&\text{feature space:} && f(x)\\
&\text{classifier space:} && Wf(x) +b\\
&\text{probability space:} && \hat{p}_i = \frac{e^{w_i f(x)} }{\sum_{j=1}^k e^{w_j f(x)}}, 
	\, i=1,\ldots,k
\label{eq:classifier}
\end{align}
where $W$, for the MSTAR classification network, is the $10\times256$ weight matrix, $b$ is the bias vector of dimension 10, and
$w_i$ is the $i^\text{th}$ row of $W$. $\hat{P}=[\hat{p}_1, ..., \hat{p}_k]^T$ is a 10-D vector in the probability space, which is directly used to determine the most likely class label (Fig. \ref{fig:intro_2}, right).

Typical deep learning networks only use the classification loss and suffer overfitting problems (Fig. \ref{fig:main_results}). We propose to alleviate this issue by applying a dual loss criterion, which augments the classification loss with an embedding loss.

\subsection{Dual loss}

Sun \etal \cite{sun2014deep} have shown that
person re-identification is better achieved by simultaneously using both face identification and verification signals as supervision (we referred to ``dual loss'' hereafter), one for recognizing the identity of a person (which is available for training images), and the other for learning feature representations that map images of the same person closer than those of different persons. The face identification task increases the
inter-personal variations by drawing features from different identities
apart, while the face verification task reduces the intra-personal variations by
pulling features from the same identity together. 
Both tasks are essential to face recognition and best achieved when images are tightly clustered within the same identities and well separated between different identities.
At the test phase, one can infer whether two new face images belong to the same, albeit {\it unknown}, identity.

We adopt the same dual loss idea for the classification task, i.e. to classify whether a new image belongs to one of the classes seen during training. Specifically, we combine classification and embedding losses to classify aerial images. The dual loss leads to better generalization than the single classification loss, because the embedding loss enforces larger margins between classes.

Our dual loss is formulated as follows:
\begin{equation}
\mathcal{L}=\mathcal{L}_{classification}+\lambda \mathcal{L}_{embedding}+R(W)
\end{equation}
where $\mathcal{L}_{classification}$ is the standard softmax cross-entropy loss, $\mathcal{L}_{embedding}$ is an embedding loss, $\lambda$ is a hyper-parameter that controls the relative importance of the embedding loss, and $R(W)$ is the standard weight decay regularization term.

Minimizing the classification loss forces the network to learn the correct predictions on individual images, whereas minimizing the embedding loss forces the network to develop an intermediate representation with  better cohesion within each class and/or  better separation between classes.  

We consider two embedding loss functions, contrastive loss and center loss. They both help learn more discriminative embeddings: contrastive loss pulls images in the same class closer and pushes those from different classes apart, whereas center loss decreases the variation of each individual classes.




{\bf Contrastive Embedding Loss}
Consider a pair of images $(x_a, x_b)$ and their associated labels $(y_a, y_b)$. The goal is to pull the network activation vectors $f(x_a)$ and $f(x_b)$ closer if they belong to the same class and push them apart if they belong to different classes.
The contrastive loss of a pair of images is:
\begin{align}
\mathcal{L}_{contrastive} &=
    \begin{cases}
        \mathcal{L}_{sim}, & \text{if } y_a = y_b \\
        \mathcal{L}_{dis}, & \text{if } y_a \neq y_b
    \end{cases} \\
\mathcal{L}_{sim} &= \max(0,\|f(x_a)-f(x_b)\|_2^2-m_s) \\
\mathcal{L}_{dis} &= \max(0,m_d-\|f(x_a)-f(x_b)\|_2^2)
\end{align}
where $f(x)$ is the network output for the layer upon which the contrastive loss is applied; and $m_s$ and $m_d$ are hyper-parameters for similar and dissimilar margins, respectively.
For images of the same label, $\mathcal{L}_{sim}>0$ only if $\|f(x_a)-f(x_b)\|_2^2>m_s$, whereas for images of different labels,  $\mathcal{L}_{dis}>0$ only if $\|f(x_a)-f(x_b)\|_2^2<m_d$.  That is, the contrastive loss acts to pull the images of the same label closer than $m_s$, and to push the images of different labels farther apart than $m_d$.

We set $m_s = 0$ in experiments for simplicity. Details of the influence of  hyper-parameters $m_d$ and $\lambda$ could be found in Section \ref{sec:ablation}.
As shown in Fig. \ref{fig:siamese}, the network with contrastive loss is implemented as a Siamese network \cite{bromley1994signature}, where a pair of  image chips  is fed in parallel to two identical networks with shared weights. Please refer to Section \ref{sec:architecture} for training details.

{\bf Center Embedding Loss.}
First proposed by Wen \etal to learn discriminative feature representations for face recognition \cite{wen2016discriminative}, 
the center loss function reduces the variance of feature vectors with the same class label, promoting tighter class clusters in the embedding representation. The center loss is applied to a minibatch of training samples as follows:
\begin{equation}
\label{eq:center}
\mathcal{L}_{center} = \frac{1}{2B} \sum_{i=1}^{B}\| f(x_i)-c_{y_i}\|_2^2
\end{equation}
where $B$ is the batch size; $x_i$ is the $i^\text{th}$ image in the batch; $f(x)$ is the network output for the layer upon which the embedding loss is applied; $y_i$ is the label of $x_i$; and $c_{y}$ is the mean of all $f(x)$ vectors belonging to class $y$. Details for choosing hyper-parameters of center loss could be found in Section \ref{sec:ablation}.

The embedding loss is computed over a batch of images in the training phase. For example, contrastive embedding loss functions on a pair of images in the batch; center embedding loss requires the mean value of feature vectors $f(x)$ of images in the batch. In the validation or test phase, the network forwards images one by one and predicts its corresponding class, as no loss needs to be computed. Therefore, the embedding loss is only computed in training phase and will not incur any additional computational burden when deployed.

\subsection{Feature Space vs. Classifier Space}
\label{sec:space}

We investigate which network layers the embedding loss should apply to. 

For the Re-ID task that employs both classification and embedding losses  \cite{sun2014deep, wen2016discriminative, schroff2015facenet}, the classifier space is used only during training when there is a finite number of \textit{known} person identities. At the test time, the classifier layer is discarded, and the network relies on the feature space only to determine if two images come from the same or different (most likely never seen) identities.

For our classification task, however, the classifier space is used in both training and testing phase.  While having a discriminative feature space for the classification task is desirable, it is ultimately the classifier and probability spaces where we need strong separation between classes.  We have the option of applying the embedding loss to the feature space, the classifier space, or the probability space.
Our hypothesis is that moving the embedding loss from the feature space to the classifier space directly will  improve classification performance the most at the test phase.


In the following sections, we conduct extensive experiments to evaluate the performance of the proposed method and hypothesis on aerial image classification.

\section{MSTAR Target Classification}
\label{sec:class}


We conduct SAR-ATR experiments to evaluate the proposed method. In addition, we visualize how applying the embedding loss to different spaces affects the separation of classes in the classifier space and the generalization capability of trained neural networks.

\subsection{Dataset}

\begin{table*}[ht]
\centering
\caption{Setup of the MSTAR dataset (augmented version of \cite{ding}) }
\label{setup}
\begin{tabular}{|c|c|c|c|c|c|c|c|c|c|c|c|c|c|c|}
\hline
\multirow{2}{*}{\textbf{Dataset}} & \multicolumn{3}{c|}{\textbf{BMP2}} & \textbf{BTR70} & \multicolumn{3}{c|}{\textbf{T72}} & \textbf{BTR60} & \textbf{2S1} & \textbf{BRDM2} & \textbf{D7}  & \textbf{T62} & \textbf{ZIL131} & \textbf{ZSU23/4} \\ \cline{2-15} 
                             & snc21  & sn9563  & sn9566 & c71   & sn132   & sn812  & sns7  &       &     &       &     &     &        &         \\ \hline
Training                     & 2563    &         &        & 2563   & 2552     &        &       & 2816   & 3289 & 3278   & 3289 & 3289 & 3289    & 3289     \\ \hline
Testing                      & 2156    & 2145     & 2156    & 2156   & 2156     & 2145    & 2101   & 2145   & 3014 & 3014   & 3014 & 3003 & 3014    & 3014     \\ \hline
\end{tabular}
\end{table*}

We evaluate the performance of the proposed method on MSTAR dataset \cite{chen7}. The dataset consists of X-band SAR image chips with 0.3m $\times$ 0.3m resolution of ten target classes, including BMP2 (infantry combat vehicle), BTR70 (armored personnel carrier), T72 (tank), etc. All chips are of size 128$ \times $128. We preprocess the chips by taking the square root of the magnitude of the original complex SAR data.

We use the same training-testing data split as Ding \etal \cite{ding}, where the training images
are obtained at a depression angle of 17\degree, and the testing images at 15\degree.
Table \ref{setup} lists 
the distribution of images by class (after translation augmentation). We randomly partition the training set, using 90\% of the data for training and 10\% for validation.
 
We split the test data into two separate sets to evaluate the performance of the proposed methods separately on \textit{seen} and \textit{unseen} target serial numbers. The \textit{unseen} test set consists of targets with serial numbers never shown to the network in the training phase, specifically BMP2 serial numbers sn9563 and sn9566 and T72 serial numbers sn812 and sns7. The \textit{seen} test set contains the rest of the test samples. 

To allow for translation variance and to make the MSTAR classification task more challenging and closer to the real-world setting, we follow the data translation augmentation method from Ding \etal \cite{ding}.
Specifically, we augment both the training and test data with ten random translated copies of each chip, mirroring the pixels at the edge when needed.

\subsection{Network architecture}
\label{sec:architecture}

\begin{figure*}[t!]
\centering
\includegraphics[width=0.98\textwidth]{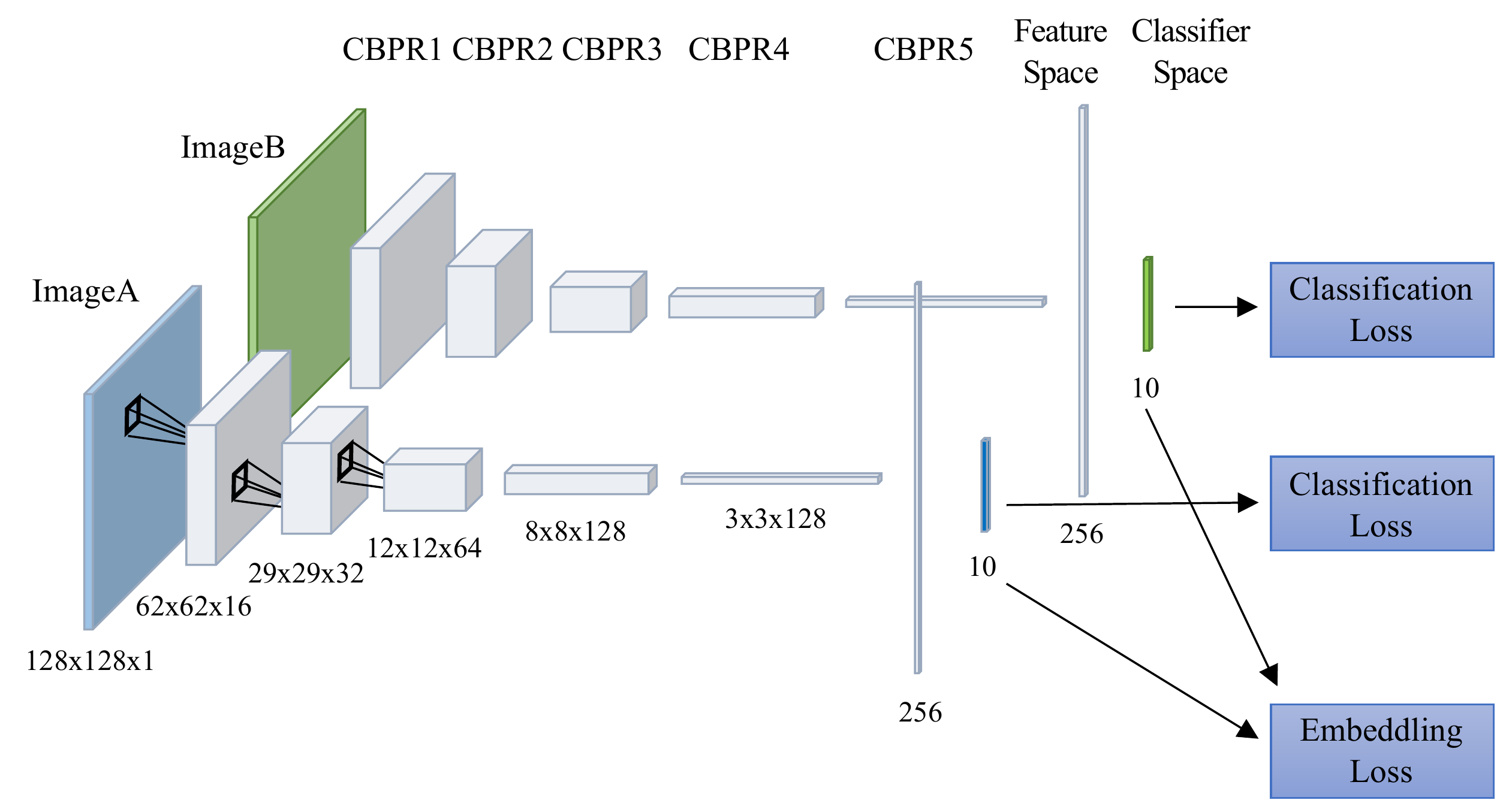}
\caption{Siamese network architecture. For a pair of images, the contrastive loss aims to learn better embeddings: pull the vectors output from the network closer if they belong to the same class and push them apart if they belong to different classes. Classification loss aims to discriminate different classes. All weights for both image A and image B are shared.}
 \label{fig:siamese}
\end{figure*}

Our CNN architecture (Fig. \ref{fig:siamese}) is based upon Chen's A-ConvNet \cite{chen} with the following modifications.

\begin{enumerate}
   \item We insert an additional convolution block to handle input chips of size $128 \times 128$, rather than cropping the input of size $88 \times 88$.
   
   \item  We add an additional fully connected layer to create an explicit feature vector of dimension $256$.
   
   \item We add batch normalization \cite{ioffe2015batch} before the ReLU layers.
\end{enumerate}

The resulting network consists of five convolutional blocks and two fully connected blocks.
Each convolutional block, or ``CBRP'' block, is a series of layers consisting of convolution, batch normalization, ReLU activation, and then pooling. The fourth block is an exception, as it does not have a pooling layer. The five convolutional layers have square kernels of sizes $5$, $5$, $6$, $5$ and $3$. The pooling layers all use kernel size $2 \times 2$. The number of output channels for the five convolutions are $16$, $32$, $64$, $128$ and $128$. The output size of each block are $62\times62\times 16, 29\times29\times 32, 12\times12\times 64, 8\times8\times 128, 3\times3\times 128$.

After the convolutional blocks, we use a fully connected layer followed by a ReLU layer to produce a $256$ dimensional feature vector. 
While each CNN layer produces an intermediate feature representation of the input image, 
 we refer to this last ${256}$-dimensional space in which this vector lies in as the \textit{feature space}.
We use a fully connected layer to learn a linear transform from the feature space to a $k$-dimensional \textit{classifier space}, where $k$ is the total number of classes. For the MSTAR dataset, $k=10$.

Fig. \ref{fig:siamese} illustrates the architecture as a Siamese network, sharing weights between two parallel networks. In practice, chips are passed through one single network in minibatches. For contrastive loss, the minibatches for ImageA and ImageB are concatenated, embedded by the network, and eventually passed to the contrastive embedding loss, which converts the combined minibatch back into pairs. When the embedding loss is center loss, the same single network is used and the center loss operates across the minibatch.

\subsection{Methods}
We evaluate the performance of our methods on the 10-class target classification problem. Our baseline network uses the standard classification loss (softmax cross-entropy loss) only,  with the network architecture described in Section \ref{sec:architecture}. For the sake of comparison, we also implement and test the network architecture from \cite{ding} and \cite{spp-sar}. We experiment with the following three possible dual loss  modifications to our baseline approach with a single classification loss:
\begin{enumerate}
\item Add the embedding loss (center or contrastive loss) to the feature space, following the previous face Re-ID work \cite{sun2014deep,wen2016discriminative}. 
\item Add the embedding loss to classifier space, which is unlike the previous face ReID work.
\item Add the embedding loss to the probability space, in attempt to move the embedding loss even further down the network.
\end{enumerate} 

\subsection{Results}
 Table \ref{result} lists
the classification accuracy of all the methods. Note that we conduct 5-fold evaluation and report the standard deviation of the performance. Fig. \ref{confmat} shows the confusion matrices of contrastive loss. We summarize the experimental results as follows:

   \begin{figure*}[htb]
\centering
\includegraphics[width=0.95\textwidth]{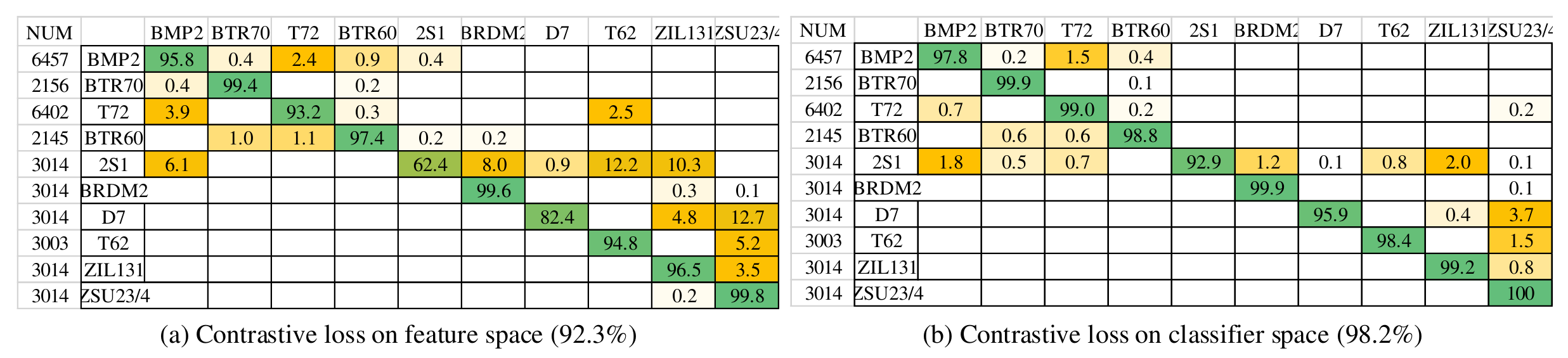}
\caption{Confusion matrix of different methods. (a) Confusion matrix of contrastive loss on feature space, the inter-class confusion is common; (b) Confusion matrix of contrastive loss on classifier space, the inter-class confusion is alleviated. }
 \label{confmat}
\end{figure*}

\begin{table}[ht]
\centering
\caption{Classification accuracy(\%) on translation augmented test set }
\label{result}
\scalebox{0.93}{\begin{tabular}{|l|c|c|c|}
\hline
\textbf{Network} & \textbf{Seen} & \textbf{Unseen} & \textbf{Overall} \\ \hline
Ding \cite{ding} & 83.0 & 80.4 & 82.4$\pm$0.13  \\ \hline
SPP \cite{spp-sar} & 90.5 & 84.1 & 88.9$\pm$0.11 \\ \hline
Baseline: classification loss only & 91.6 & 91.0 & 91.4$\pm$0.12 \\ \hline
Baseline + center loss on feature space & 97.4 & 92.7 & 96.3$\pm$0.10 \\ \hline
Baseline + contrastive loss on feature space & 92.2 & 92.6 & 92.3$\pm$0.11 \\ \hline
Baseline + center loss on classifier space  & \textbf{98.7} & 96.7 & \textbf{98.2}$\pm$0.09 \\ \hline
Baseline + contrastive loss on classifier space & \textbf{98.5} & \textbf{97.2} & \textbf{98.2}$\pm$0.10 \\ \hline
Baseline + contrastive loss on probability space & 90.1 & 92.4 & 90.6$\pm$0.13 \\ \hline

\end{tabular}
}
\end{table}

\begin{enumerate}
\setlength{\itemsep}{3pt}
\item \textbf{Our baseline is better than previous networks}. Compared to Ding \etal \cite{ding}, the overall accuracy of the baseline improves 9.0\%, and the unseen set result improves 10.6\%. Compared to SPP network \cite{spp-sar}, the overall accuracy improves 2.5\%, and the unseen set result improves 6.9\%. Note that we adopt the same translation methods in the test phase as in \cite{ding}, which is a much more difficult setting than testing the original untranslated chip set.

\item
\textbf{Adding an embedding loss in the feature space improves test classification accuracy.} Compared with our baseline, where only the classification loss is used, contrastive loss and center loss both improve the classification accuracy on the test set. For the center loss, compared with baseline, applying the embedding loss on feature space improves the overall classification accuracy by 4.9\% and the unseen classification accuracy by 1.7\%. For the contrastive loss, compared with the baseline, applying the embedding loss on feature space improves the overall classification accuracy by 0.9\% and the unseen classification accuracy by 1.6\%.

\item
\textbf{Moving the embedding loss from the feature space to the classifier space further improves the test accuracy.} For the center loss, moving from feature space to classifier space further improves overall classification accuracy by 1.9\% and the unseen classification accuracy by 4.0\%. 
When moving contrastive loss from feature space to classifier space, we further improve the overall classification accuracy by 5.9\% and unseen classification accuracy by 4.6\%. On the unseen test set, the influence of applying embedding loss on classifier space is more significant.


\item
\textbf{Moving the embedding loss from the classifier space to the probability space hurts the classification accuracy.} When applying the contrastive loss after the softmax non-linear layer, the overall classification accuracy decreases 7.6\% and the unseen classification accuracy decreases 4.8\%. This demonstrates that adding embedding loss to the probability space shall not help the classification performance. 
\end{enumerate}

Overall, adding a feature embedding loss improves the classification accuracy at the test time.  Applying the embedding loss in the classifier space leads to higher overall classification accuracy than that applying it in the feature space. Additionally, although both embedding loss functions perform equivalently well, the contrastive loss performs slightly better than the center loss on unseen test images.

\subsection{Analysis}
In order to better understand how the proposed dual loss influences the learned space for target classification, we visualize the learned 10-D classifier space in 2-D using both orthogonal projection and t-Distributed Stochastic Neighbor Embedding (t-SNE) \cite{tsne}. 

\textbf{Visualization of two confusing classes.}
Fig. \ref{fig:ortho} plots the representation in the classifier space for two specific target classes, 2S1 and BRDM2. 
We focus on the distribution of the test points that are never shown to the network (dots in Fig. \ref{fig:ortho}). 
With the classification loss only, the intra-class variance is visibly large (Fig. \ref{fig:ortho} left). Although the inter-class overlapping of training data is not common, due to high intra-class variance, training with classification loss only still confuses the two classes 8.6\% of the test data points. When the embedding loss is applied to the feature space, the intra-class variance is visibly decreased (Fig. \ref{fig:ortho} center).  However, the difference between the two classes is still small, leading to 8.0\% inter-class confusion. When we move the embedding loss to the classifier space, the intra-class variance is further decreased, and the inter-class separation is more significant (Fig. \ref{fig:ortho} right). Furthermore, the distribution of the two classes is more aligned with their respective axes, allowing the network to be more confident of its classification and reducing the confusion between 2S1 and BRDM2 to only 1.2\%.

\def\widthA{0.4}
\def\widthB{0}
\def\widthC{0.33}

\begin{figure*}[t]
\centering
\begin{tabular}{c|@{}m{0.4\linewidth}@{}|@{}m{0.4\linewidth}@{}|}
     \multicolumn{1}{c}{} &
     \multicolumn{1}{c}{Embedding loss on feature space} &
     \multicolumn{1}{c}{Embedding loss on classifier space} \\
     \hhline{~|-|-|}
     \rotatebox[origin=c]{90}{Contrastive loss} &
     \imwX{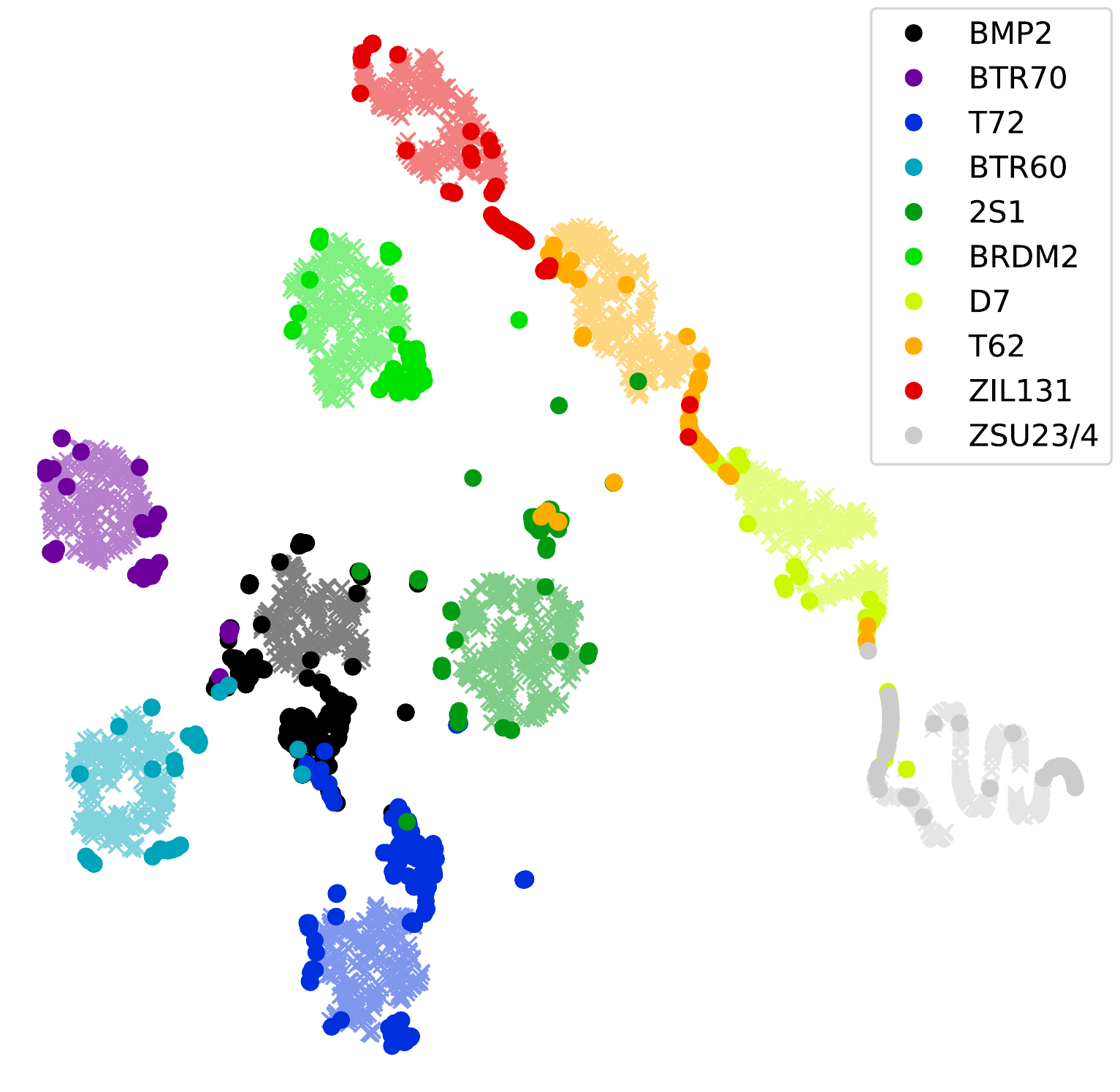}{\widthA} &
     \imwX{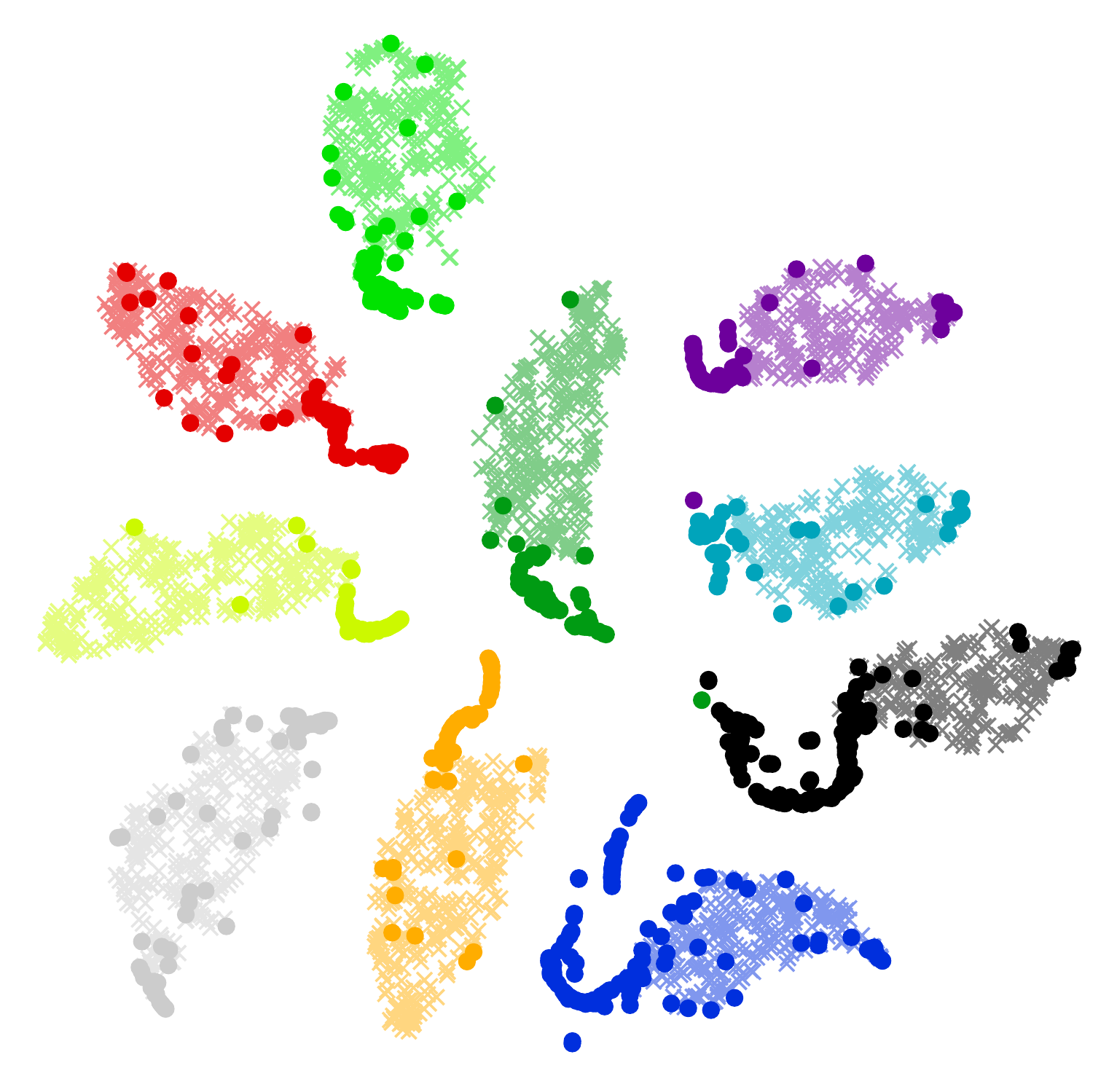}{\widthA} \\
     \hhline{~|-|-|}
     \rotatebox[origin=c]{90}{Center loss} &
     \imwX{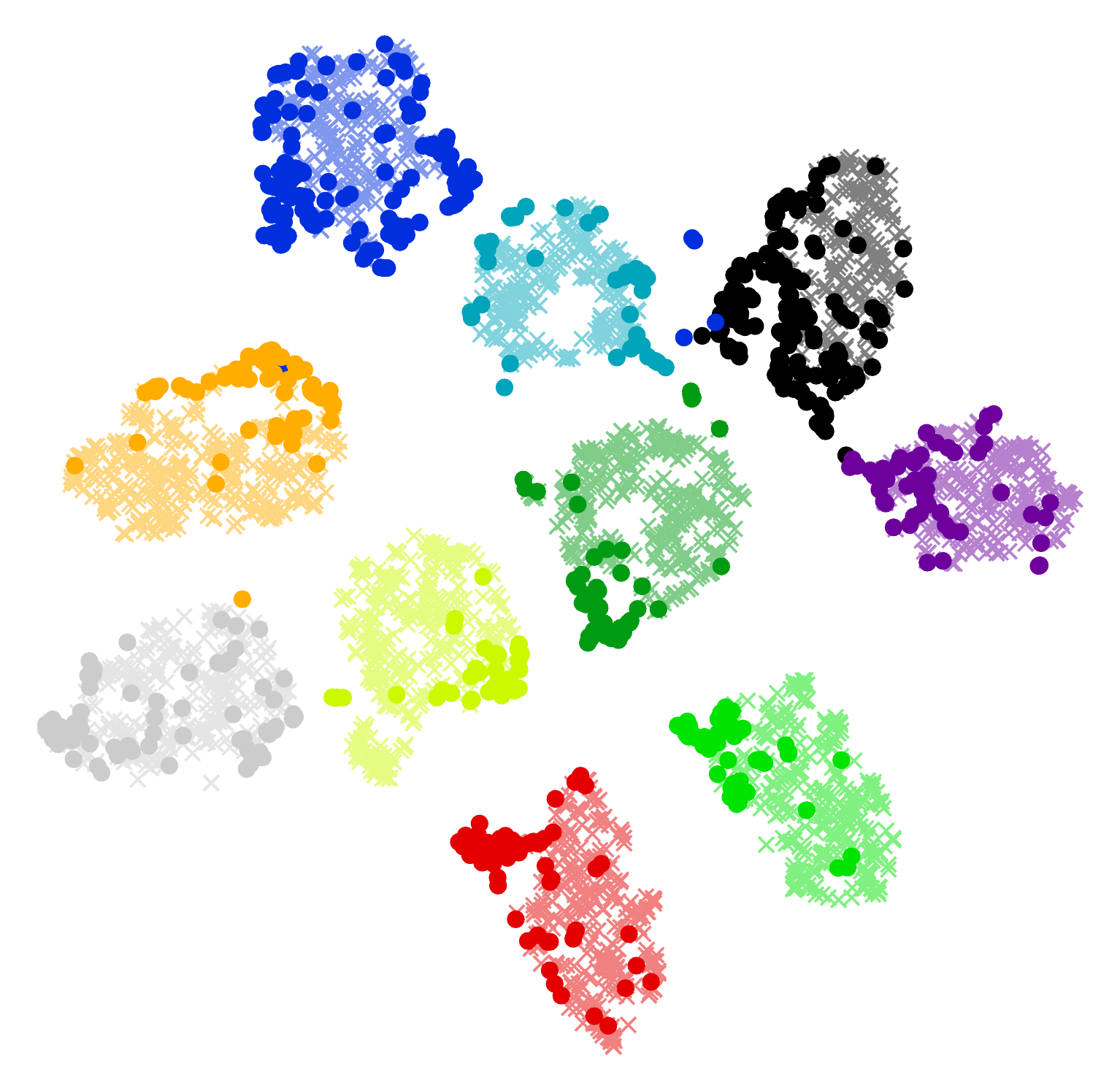}{\widthA} &
     \imwX{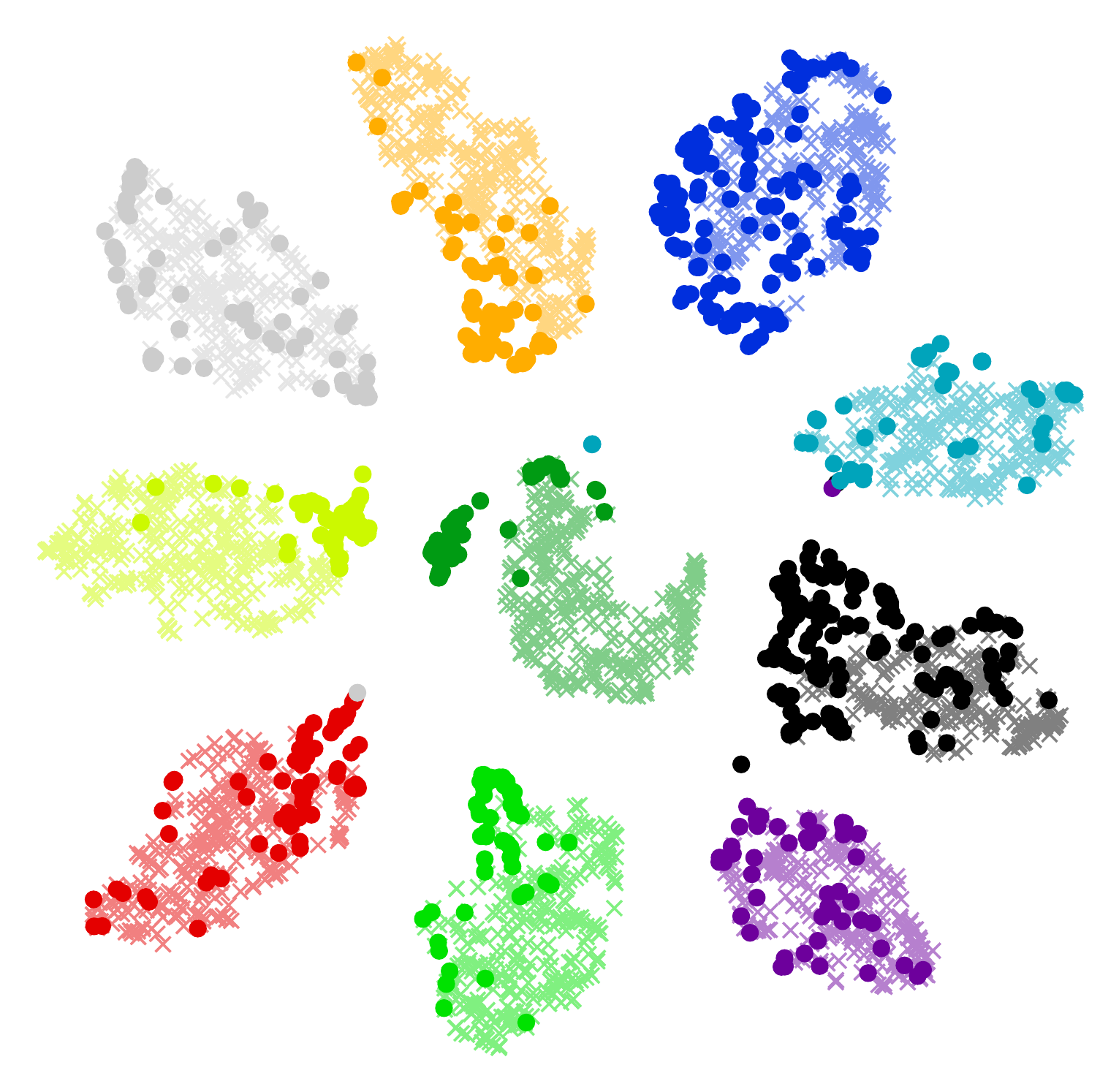}{\widthA} \\
     \hhline{~|-|-|}
\end{tabular}
\caption{t-SNE visualization of the 10-D classifier space projected onto a 2D space, comparing contrastive loss (top row) to center loss (bottom row) and comparing the loss applied to the feature space (left column) and to the classifier space (right column). When the contrastive loss is moved from the feature space (top, left) to classifier space (top, right), the learned classifier space become noticeably more discriminative, i.e smaller intra-class variance and bigger inter-class separation. Although the center loss applied to the feature space (bottom, left) produces reasonable clustering in the classifier space, we observe more discriminative embedding  when moving the center loss to the classifier space (bottom, right).}
\label{fig:tsne}
\end{figure*}

\begin{figure*}[htb]
\centering
\includegraphics[width=0.95\textwidth]{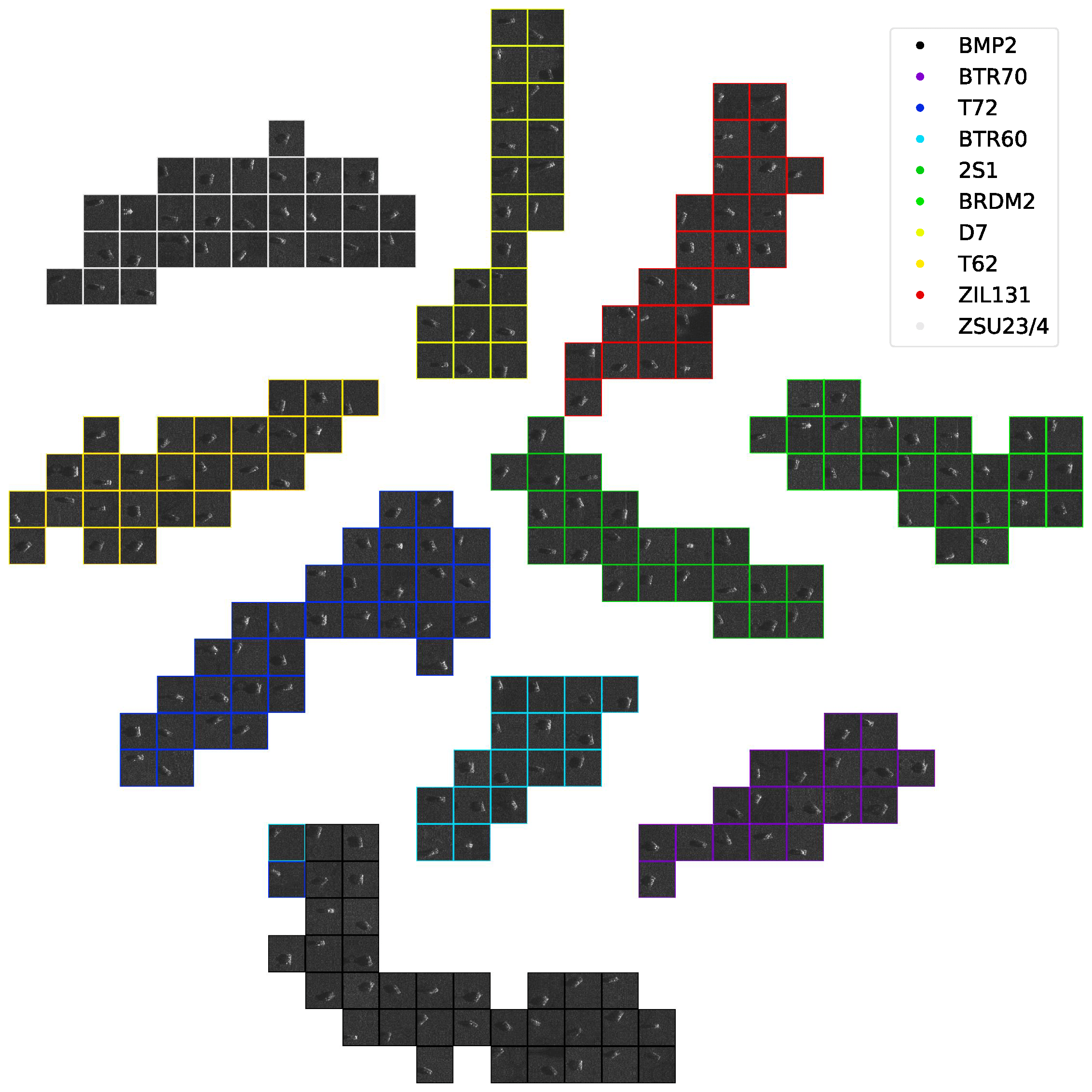}
\caption{ t-SNE visualization of the embeddings in the classifier space of the CNN trained with the contrastive loss on the classifier space. Data points are represented by their associated chip to show the distribution of chips in the embedded classifier space. The plot shows that the proposed method is invariant to both translation and vehicle orientation. }
 \label{embedding}
\end{figure*}



\textbf{t-SNE visualization.}
t-SNE \cite{tsne} technique can visualize high-dimensional data by mapping each data point to location in a two or three-dimensional space. 
In Fig. \ref{fig:tsne}, we visualize the classifier space learned by applying the embedding loss in different spaces. The quality of the class clusters in this t-SNE visualization mirrors the quantitative accuracy results in Table \ref{result}.
Moving the contrastive embedding loss from the feature space to the classifier space, the cluster of each class has smaller intra-class spread and bigger inter-class distance. In addition, the number of misclassified points in the test set are reduced when we apply the embedding loss to the classifier layer. t-SNE visualization shows that the center loss applied to the feature space induces good clustering in the classifier space.  

Fig. \ref{embedding} shows the t-SNE embedding of the contrastive loss applied to the classifier space, with each test point represented by its associated chip. This visualization shows a uniform distribution of both vehicle rotation and position within the chip, indicating that the trained network is invariant to both translation and target orientation.

\begin{figure*}[ht]
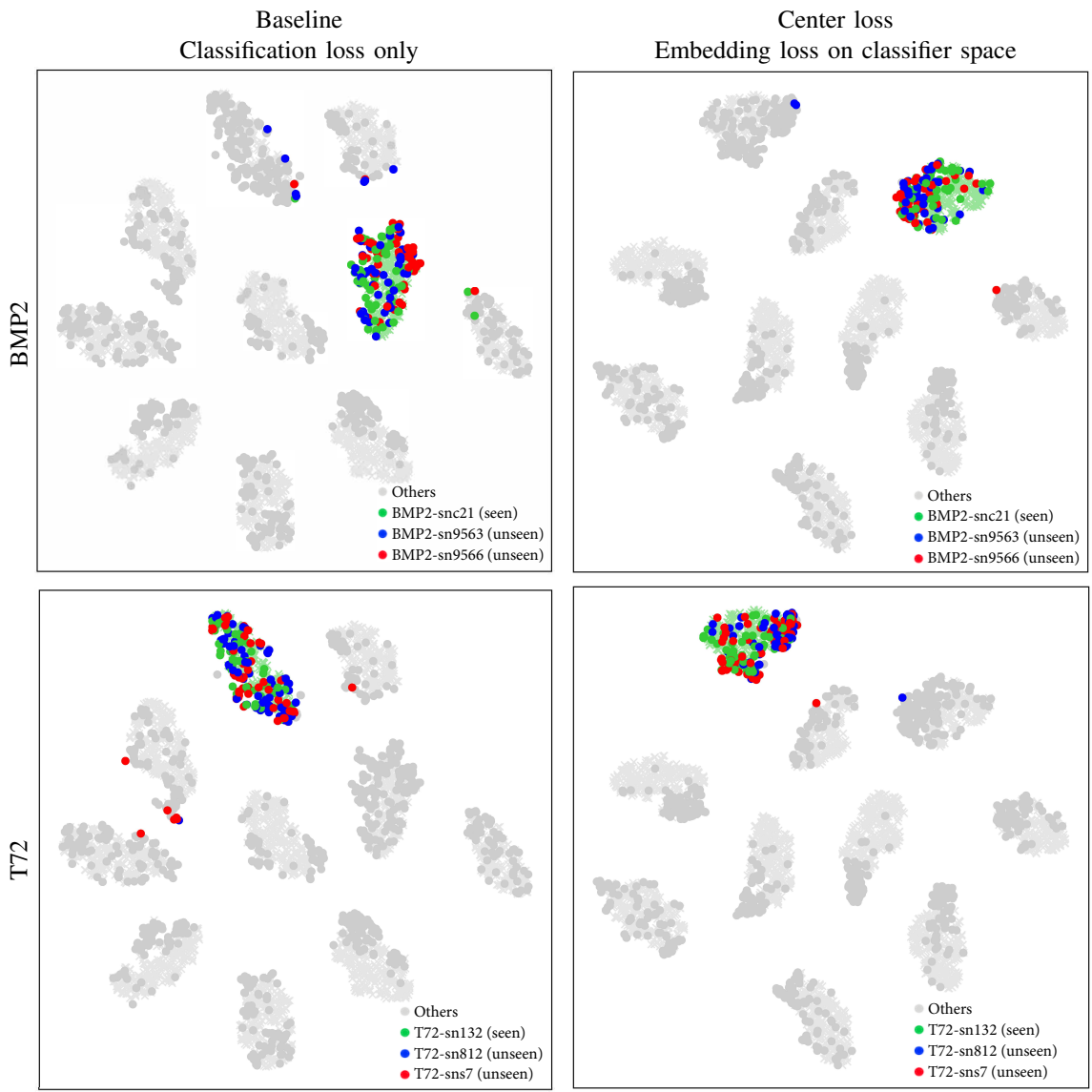

\centering
\tb{@{}cccc@{}}{0.5}{
& {Baseline} & Center loss \\
& {Classification loss only} &  Embedding loss on classifier space \\
	\multirow{1}{*}[22.4ex]{\rotatebox{90}{BMP2}}  & \imw__{baselinea}{\widthA}{\footnotesize } &
	\imw__{centera}{\widthA}{\footnotesize } \\
	\multirow{1}{*}[20.4ex]{\rotatebox[origin=c]{90}{T72}} &\imw__{baselineb}{\widthA}{\footnotesize } &
	\imw__{centerb}{\widthA}{\footnotesize } 
}
\caption{
Dual loss could decrease the training and testing variations.
Moving from classification loss to dual loss, targets with unseen serial numbers are more often embedded near the training instances, leading to an 6\% improvement in unseen classification accuracy. t-SNE visualization on the classifier space, shows how dual loss can be more robust to variations between training and test sets. Bold dots represent a subset of test points, while faint X’s represent a subset of training points. Green indicates targets with a serial number seen in the training set, while blue and red points represent test targets with serial numbers unseen during training time. The BMP2 and T72 targets are highlighted in the top and bottom rows, respectively, while all other classes are shown in gray for reference.
}
\label{fig:seen_unseen}
\end{figure*}

The proposed dual loss allows the network to become more robust to variations between training and test sets. As depicted in Fig. \ref{fig:seen_unseen}, training with the classification loss only, there were a number of poorly embedded BMP2 and T72 targets, specially for unseen serial numbers. The variation causes low recognition accuracy.
Trained with the dual loss on classifier space, the difference between embedded training and test samples are relieved, and the recognition accuracy has been improved.

\subsection{Influence of hyper-parameters and network architecture}
To understand how different component of the proposed method affect the network performance, we conduct experiments varying hyper-parameters of dual loss functions and network architectures.
\label{sec:ablation}
\subsubsection{Contrastive Embedding Loss}
We focus on two hyper-parameters in the contrastive loss function (we set the third hyper-parameter similar margin $m_s$ to 0 for simplicity): dissimilar margin $m_d$ which controls the distance of negative pairs, and embedding loss weight $\lambda$ which controls the emphasis on contrastive embedding loss. Table \ref{tab:contra_1} compares the contrastive loss with varying $m_d$ at $\lambda = 1.0$. Table \ref{tab:contra_2} compares the contrastive loss with varying $\lambda$ at $m_d = 1.0$. The contrastive loss on classifier space generally performs better than that on the feature space. When fixing $\lambda = 1.0$, the contrastive loss on feature space achieves highest classification accuracy $88.6 \%$ at $m_d = 0.8$, while the contrastive loss on classifier space achieves highest classification accuracy $98.2 \%$ at $m_d = 1.0$. When fixing $m_d = 1.0$, the contrastive loss on feature space achieves highest classification accuracy $92.3 \%$ at $\lambda = 0.15$, while the contrastive loss on classifier space achieves highest classification accuracy $98.2 \%$ at $\lambda = 1.0$.

\begin{table}[]
\centering
\caption{Classification accuracy(\%) of contrastive loss with varying $m_d$ and fixed $\lambda$ = 1.0 (\%)}
\label{tab:contra_1}
\begin{tabular}{|c|c|c|c|}
\hline
\backslashbox{Loss type} {$m_d$}                               & 0.8  & 1.0  & 1.2  \\ \hline
contrastive loss on feature space    & 88.6 & 86.3 & 87.7 \\ \hline
contrastive loss on classifier space & 97.6 & \textbf{98.2} & 98.0 \\ \hline
\end{tabular}
\end{table}

\begin{table}[]
\centering
\caption{Classification accuracy(\%) of contrastive loss with varying $\lambda$ and fixed $m_d$ = 1.0}
\label{tab:contra_2}
\begin{tabular}{|c|c|c|c|c|}
\hline
\backslashbox{Loss type} {$\lambda$}                               & 0.15  & 0.3  & 0.8 & 1.0  \\ \hline
contrastive loss on feature space    & 92.3 & 92.1 & 87.5 & 86.3 \\ \hline
contrastive loss on classifier space & 96.0 & 96.5 & 97.1 & \textbf{98.2} \\ \hline
\end{tabular}
\end{table}

\subsubsection{Center Embedding Loss}

We consider hyper-parameter $\lambda$, the weight of center loss. Table \ref{tab:center_1} compares the center loss with varying $\lambda = 1.0$. The center loss on classifier space generally performs better than that on the feature space. They both achieve highest classification accuracy at $\lambda = 3\times10^{-3}$. The center loss on feature space achieves 96.3\% classification accuracy, while the center loss on classifier space achieves 98.2\% classification accuracy.

\begin{table}[]
\centering
\caption{Classification accuracy(\%) of center loss with varying $\lambda$}
\label{tab:center_1}

\begin{tabular}{|c|c|c|c|c|}
\hline
\backslashbox{Loss type} {$\lambda$}                               & 3e-1  & 3e-2 & 3e-3 & 3e-4  \\ \hline
center loss on feature space    & 95.1 & 95.4 & 96.3 & 95.8 \\ \hline
center loss on classifier space & 96.9 & 97.3 & \textbf{98.2} & 97.6 \\ \hline
\end{tabular}
\end{table}

\begin{figure}[t!]
\centering
\includegraphics[width=0.48\textwidth]{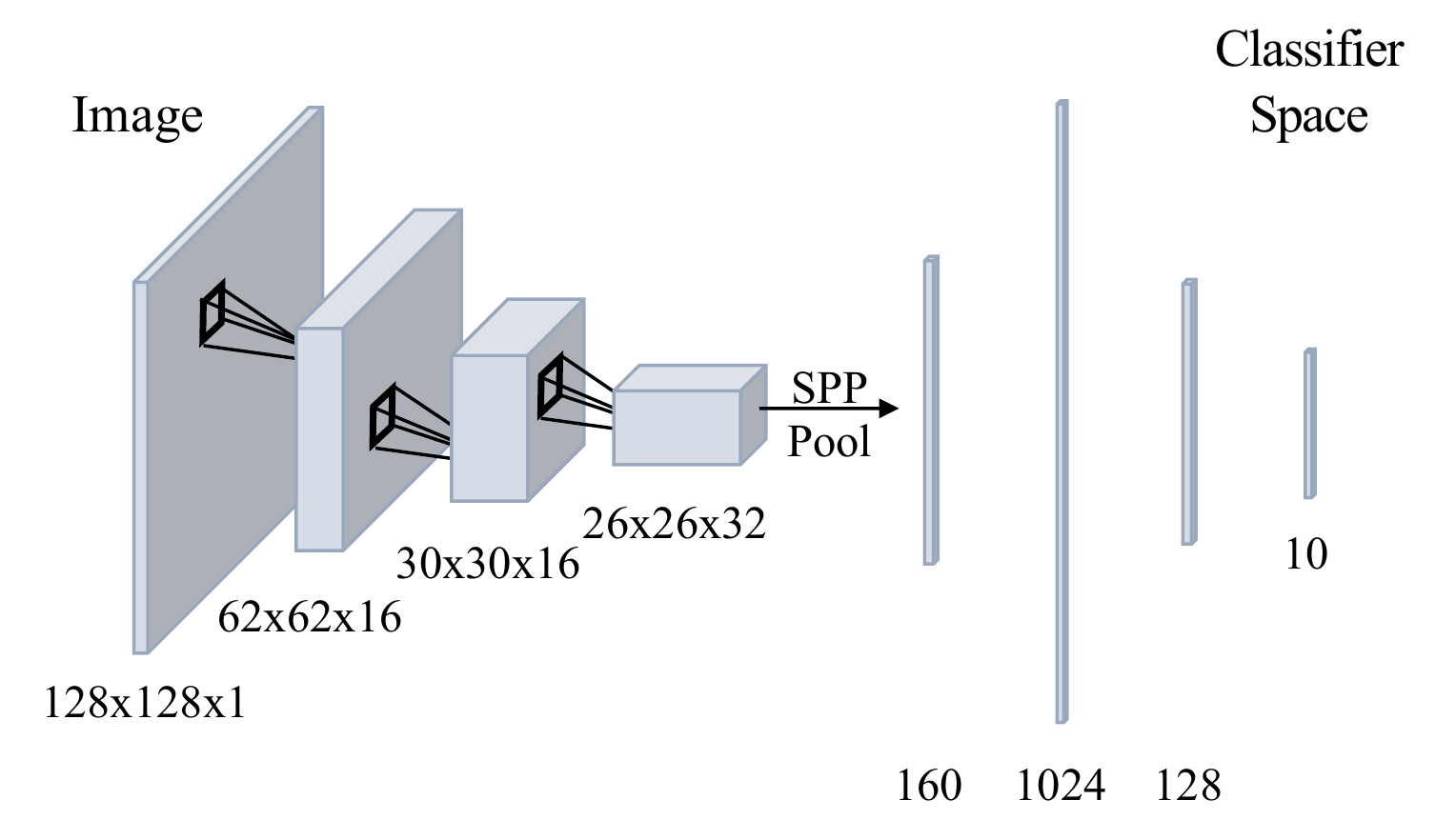}
\caption{Spatial pyramid pooling (SPP) architecture \cite{spp-sar} for SAR-ATR.}
 \label{fig:spp-arch}
\end{figure}

\subsubsection{Network architecture}

We also apply our loss functions on different network architecture to validate the robustness of the proposed method. Table \ref{tab:spp} reports the classification accuracy of SPP \cite{spp-sar} network (Fig. \ref{fig:spp-arch}). When trained with center loss on feature space, the accuracy on unseen subset improves by 0.6\%, while overall accuracy improves by 2.3\%. When trained with center loss on classifier space, the unseen accuracy improves 5.7\%, and the overall accuracy improves 5.8\%. The observation of performance gain with dual loss on classifier space does hold regardless of the changes in network architecture.

\begin{table}[ht]
\centering
\caption{Classification accuracy(\%) of different loss with SPP \cite{spp-sar}}
\label{tab:spp}
\scalebox{0.95}{\begin{tabular}{|l|c|c|c|}
\hline
\textbf{Network} & \textbf{Seen} & \textbf{Unseen} & \textbf{Overall} \\ \hline
SPP \cite{spp-sar} & 90.5 & 84.1 & 88.9$\pm$0.11 \\ \hline
SPP + center loss on feature space & 93.3 & 84.7 & 91.2$\pm$0.12 \\ \hline
SPP + contrastive loss on feature space & 92.1 & 83.2 & 89.9$\pm$0.12 \\ \hline
SPP + center loss on classifier space  & \textbf{96.2} & \textbf{89.8} & \textbf{94.7}$\pm$0.09 \\ \hline
SPP + contrastive loss on classifier space & 93.8 & 86.1 & 91.9$\pm$0.11 \\ \hline

\end{tabular}
}
\end{table}

\section{MSTAR Outlier Rejection}

\label{sec:roc}
  \begin{figure}[htb]
\centering
\includegraphics[width=0.48\textwidth]{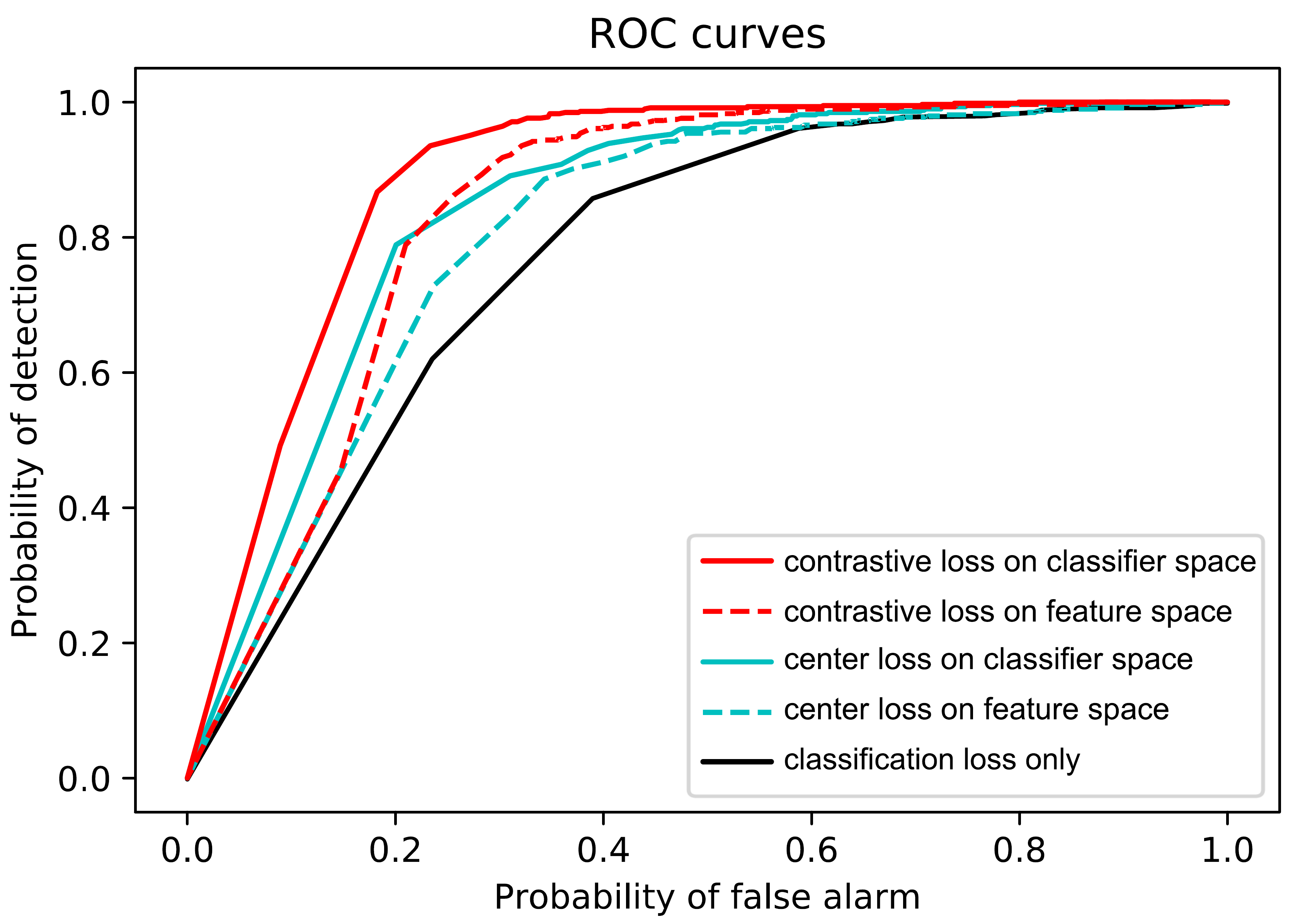}
\caption{ ROC curves (probability of detection versus that of false alarm) of different losses. Dual loss performs better than classification loss only. Applying the embedding loss to the classifier space performs better than applying it to the feature space. The contrastive loss outperforms the center loss for outlier rejection. }
 \label{ROC}
\end{figure}

In SAR-ATR systems, the detector may return some unseen targets of unknown classes which could confuse the system. We conduct outlier rejection experiments to evaluate the confuser rejection performance of the proposed methods. Note that the network architecture is the same as in Section \ref{sec:architecture}.

\subsection{Data}
We follow the same experimental setup on outlier rejection as Chen \cite{chen}, training the neural network on three target classes, BMP-2, BTR-70, and T-72, and testing on the three training targets and two additional confuser targets, 2S1 and ZIL-131. In this experimental scenario, the test set is composed of 588 images of three known targets (196 images per class) and 548 images of two confuser targets (274 images per class).

\subsection{Methods}
The output of the CNN gives the posterior probability of classification for each class. Given a threshold $\tau$, the target image will be declared as a confuser if all the posterior probabilities are lower than $\tau$. Denote $P_d$ as the probability of detection, $P_{fa}$ as the probability of false alarm. These values are formulated as follows:

\begin{equation}
P_d = \frac{N(\mbox{prediction = target}|\mbox{real label = target} )}{N(\mbox{real label = target)}}
\end{equation}

\begin{equation}
P_{fa} = \frac{N(\mbox{prediction = target}|\mbox{real label = confuser} )}{N(\mbox{real label = confuser)}}
\end{equation}

Varying $\tau$, we collect many pairs of detection probability and false alarm probability. 
The receiver operating characteristic (ROC) curves describe the relationship between $P_d$ and $P_{fa}$.

\subsection{Results}
The ROC curves of different methods are shown in Fig. \ref{ROC}.
At $P_d = 90 $\%, the false alarm probabilities 
are 56.1\%, 39.9\%, 20.5\% for
 our baseline, our center loss applied in the classifier space, and our contrastive loss applied in the classifier space, respectively. For each embedding loss, given a fixed detection probability, applying the loss in the classifier space gives a lower false alarm probability than applying that loss in the feature space.

\subsection{Analysis}
From these experimental results, we draw the following conclusions:
\begin{enumerate}
\item In terms of rejection performance, dual loss performs better than single classification loss only. Since the learned classifier space is more discriminative, the ATR system becomes more confident in rejecting outliers.
\item Applying the embedding loss in the classifier space performs better on outlier rejection than that in the feature space. Moving the embedding loss from the feature space to the classifier space can improve outlier rejection performance in addition to improving classification accuracy.
\item  Contrastive loss performs better than center loss. Center loss makes the classifier more discriminative by decreasing the intra-class variance. Contrastive loss, on the other hand, also increases the inter-class difference, which leads to improved outlier rejection performance.
\end{enumerate}

\section{AID Aerial Scene Classification}
Aerial scene classification is an important task in aerial image classification.
To evaluate the generalization ability of the proposed method, we also report the performance of the proposed dual loss on a benchmark aerial scene classification dataset, AID dataset. 

\subsection{Data}
\begin{figure}[htb]
\centering
\includegraphics[width=0.48\textwidth]{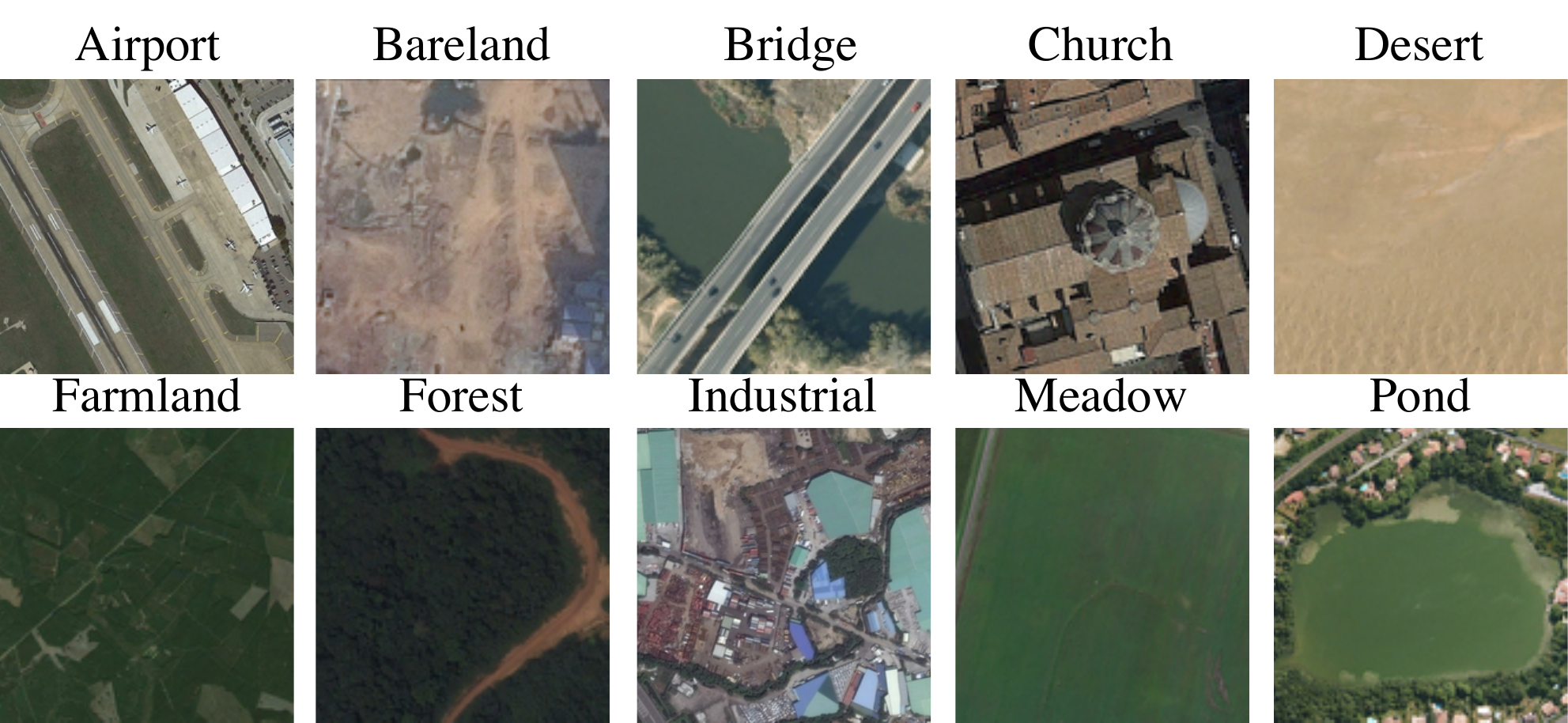}
\caption{ Some sample images from AID dataset~\cite{AID}. }
 \label{fig:aid}
\end{figure}

We conduct experiments on the AID dataset \cite{AID} (Fig. \ref{fig:aid}) to evaluate the aerial scene classification performance of the proposed method.
The dataset collects images from Google Earth imagery and consists of 30 aerial scene types, including airport, bare land, baseball field, beach, and some other common aerial scenes. The aerial images are with fixed size of $600 \times 600$ pixels
with various pixel resolutions up to half a meter. The total  images add up to 10000. \footnote{For the details of the dataset, please refer to \cite{AID}. }  We follow all the experimental settings in \cite{AID}. Specifically, we use CaffeNet \cite{ding8} as the network architecture and 50\% data for testing. For the remaining 50\% of the data, we randomly partition 90\% for training and 10\% for validation.

\subsection{Methods}

We report the recognition results of support vector machine (SVM) on pre-trained ImageNet \cite{ding8} features, as a fundamental baseline.
Similar to the Section \ref{sec:class}, we also report the aerial recognition results of classification loss only, as well as the results of our embedding loss (center loss) on feature space and classifier space on CaffeNet and VGG16 network~\cite{simonyan2014very} respectively. 

\subsection{Results and analysis}

Table \ref{tab:aid} shows the classification accuracy of different methods the on AID dataset. We conduct 5-fold evaluation and report the standard deviation of the performance. Compared to the SVM on the pre-trained model technique, end-to-end CaffeNet with classification loss improves by 3.0\%. Furthermore, when adding the center loss on classifier space, the classification accuracy is further improved by 1.5\%. Yet adding center loss on feature space leads to no performance improvement. Similarly when training with VGG16 network, adding center loss on classifier space gives 2\% gain. In summary, additional embedding loss  improves the aerial image classification performance.
\begin{table}[ht]
\centering
\caption{Recognition accuracy of different methods on AID dataset \cite{AID}}
\label{tab:aid}
\scalebox{0.95}{\begin{tabular}{|l|c|c|c|}
\hline
\textbf{Network} & \textbf{Accuracy} \\ \hline
CaffeNet + SVM & 89.5$\pm$0.12 \\ \hline
CaffeNet + classification loss only \cite{AID}  & 92.5$\pm$0.11 \\ \hline
CaffeNet + center loss on feature space & 92.3$\pm$0.11 \\ \hline
CaffeNet + center loss on classifier space  & 94.0$\pm$0.10 \\ \hline \hline
VGG16 \cite{simonyan2014very} & 95.2$\pm$0.12 \\ \hline
VGG16 + center loss on classifier space & 97.0$\pm$0.09 \\ \hline
\end{tabular}
}
\end{table}

\section{Conclusion}

\label{sec:con}

We present a dual embedding and classification loss for aerial image classification. 
We demonstrate that applying the embedding loss to the classifier space leads to better performance than applying it to the feature space. Two types of embedding loss functions, contrastive loss and center loss, are compared. 
Both can achieve higher target classification accuracy for SAR target recognition and aerial scene recognition. 
In addition, we find that contrastive loss performs better than center loss on outlier rejection tasks.

\section*{Acknowledgments}

The authors would like to thank Dr. Zhirong Wu, Baladitya Yellapragada and Michele Winter for valuable discussions and feedback.

\ifCLASSOPTIONcaptionsoff
  \newpage
\fi

\bibliographystyle{IEEEtran}
\bibliography{egbib}

\end{document}